\documentclass{article}
\usepackage{spconf,amsmath,graphicx}
\usepackage{amsmath,amsfonts}
\usepackage{algorithmic}
\usepackage{algorithm}
\usepackage{array}
\usepackage[caption=false,font=normalsize,labelfont=sf,textfont=sf]{subfig}
\usepackage{textcomp}
\usepackage{stfloats}
\usepackage{url}
\usepackage{verbatim}
\usepackage{graphicx}
\usepackage{cite}
\usepackage{amssymb}
\usepackage{units}
\usepackage{makecell}
\usepackage{multirow}
\usepackage{booktabs}

\title{EK-Net:Real-time Scene Text Detection with Expand Kernel Distance}
\name{{Boyuan Zhu}$^1$, {Fagui Liu}$^{1,2,*}$, {Xi Chen}$^{1}$, {Quan Tang}$^{1,*}$\thanks{* Corresponding author. (E-mail address: fgliu@scut.edu.cn)}\thanks{
This work was supported in part by the Guangdong Major Project of Basic and Applied Basic Research under Grant 2019B030302002, in part by the Science and Technology Major Project of Guangzhou under number 202007030006, in part by the Major Key Project of PCL, China under Grant PCL2023A09, in part by the Science and Technology Project of Guangdong Province under Grant 2021B1111600001, in part by the Engineering and Technology Research Center of Guangdong Province for Logistics Supply Chain and Internet of Things under Grant DDST[2016]176.}}
\address{$^{1}${School of Computer Science and Engineering, South China University of Technology, Guangzhou, China,} \\
$^{2}${Peng Cheng Laboratory, Shenzhen, China}}

\begin{document}
%
\maketitle
\begin{abstract}
Recently, scene text detection has received significant attention due to its wide application. However, accurate detection in complex scenes of multiple scales, orientations, and curvature remains a challenge. Numerous detection methods adopt the Vatti clipping (VC) algorithm for multiple-instance training to address the issue of arbitrary-shaped text. Yet we identify several bias results from these approaches called the \textquotedblleft shrinked kernel\textquotedblright. Specifically, it refers to a decrease in accuracy resulting from an output that overly favors the text kernel. In this paper, we propose a new approach named Expand Kernel Network (EK-Net) with expand kernel distance to compensate for the previous deficiency, which includes three-stages regression to complete instance detection. Moreover, EK-Net not only realize the precise positioning of arbitrary-shaped text, but also achieve a trade-off between performance and speed. Evaluation results demonstrate that EK-Net achieves state-of-the-art or competitive performance compared to other advanced methods, e.g., F-measure of 85.72\% at 35.42 FPS on ICDAR 2015, F-measure of 85.75\% at 40.13 FPS on CTW1500.
\end{abstract}
\section{Introduction}
\label{sec:intro}

In recent years, text spotting in scene images \cite{Zhang2021PointerNF} has acquired more popularity owning to its broad applications in automatic driving \cite{Zhou2016SemanticUO}, image and video understanding \cite{Long2018SceneTD}, and real-time translation\cite{Wang2021TowardsRV}. With the progress of CNN \cite{Jaderberg2014ReadingTI, He2016IdentityMI}, text detection obtains abundant semantic information \cite{Li2018ShapeRT,zhu2021fourier}. Researchers form lots of frameworks based on these to complete complex text detection \cite{dai2021progressive}. Nonetheless, scene text detection remains a challenging task as a result of the diverse nature of text curvatures, orientations, and aspect ratios.

\begin{figure}[t]
    \centering
        \includegraphics[width=1\columnwidth]{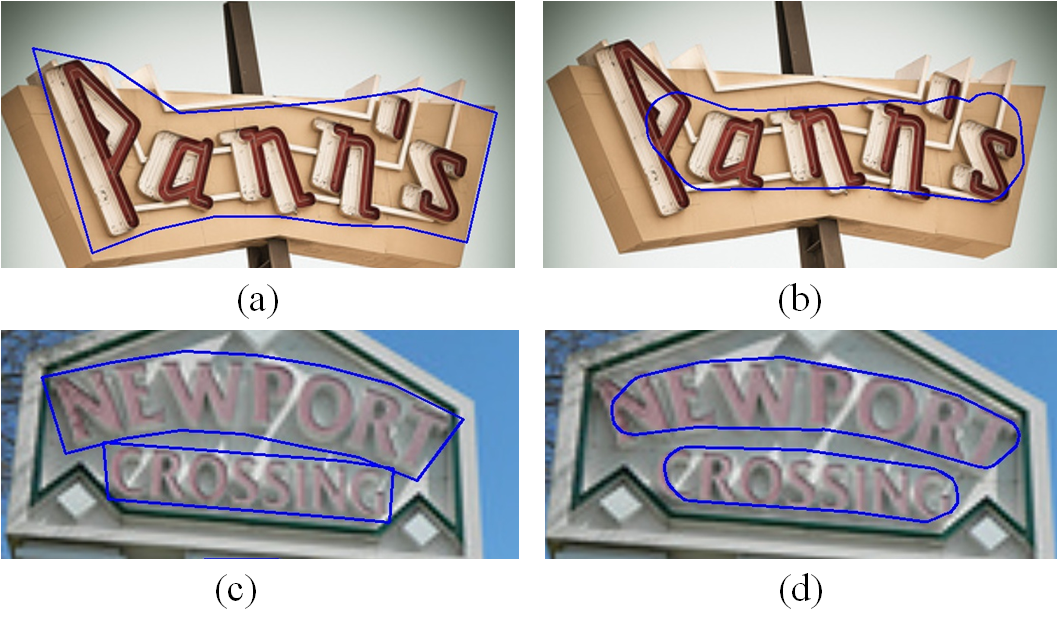}
      \caption{Limitations of kernel based methods on scene text detection. (a)(c) are GroundTruth and (b)(d) are output results. The blue lines are detection boundaries.}\label{fig:F1}
\end{figure}

Typically, two strategies are employed to address this challenge. The first approach classifies text detection into object detection domain, using rotated rectangles or quadrilaterals as text boundaries. These methods’ development at each stage is closely related to the progress of object detection. For instance, TextBoxes \cite{Liao2016TextBoxesAF} modifies the aspect ratio of the default boxes in SSD \cite{Liu2015SSDSS} to adapt to the text length. \cite{liao2018textboxes++} utilizes quadrilateral sliding windows through the DMPNet \cite{liu2017deep} in specific convolution layers to capture text regions with higher overlap. SPCnet \cite{xie2019scene} is designed by Mask R-CNN \cite{he2017mask}, which mitigates false positives by means of semantic segmentation guidance. Nonetheless, this strategy has evident deficiencies. To begin with, it overlooks the distinctive characteristics of text shapes. Text typically occupies a relatively small area within the entire image. Although the text boxes are more regular, they often possess extreme aspect ratios compared with other detection targets, and furthermore, formulating suitable anchors to accommodate texts of varying shapes is also a difficult task. 

An alternative strategy involves decomposing text instances into multiple theoretical components, which followed by the reconstruction of polygonal contours through a series of post-processing steps. PSENet \cite{Li2018ShapeRT} firstly uses Vatti clipping algorithm \cite{Vatti1992AGS} to get text kernels, which employs a progressive scale expansion algorithm to gradually expand the detected areas from small kernels to complete instances. Since then, a large number of kernel-based methods have been proposed, such as DB-Net \cite{liao2020real} and DB-Net++ \cite{Liao2019RealTimeST} combine a learnable Differentiable Binarization with kernels to achieve text instances detection. PAN \cite{Wang2019EfficientAA} and PAN++ \cite{wang2021pan++} introduce a low computational-cost segmentation head and a learnable post-processing technique based on PSENet \cite{Li2018ShapeRT}. Centripetal Text \cite{sheng2021centripetaltext} divide a text instance into a kernel with Centripetal distance. BipNet \cite{Yang2021BiPNetBP} proposes a Center Points to participate in prediction. These methods are more flexible and precise than the first approach in modeling, thus making them the mainstream strategy in text detection task. However, they have a common characteristic that leads to the defect, which we refer to \textquotedblleft shrinked kernel\textquotedblright is shown in Fig\ref{fig:F1}. Owing to text instances are decomposed into multiple parts and the kernel components that hold significant importance, these models exhibit a pronounced bias towards the center, leading to a accuracy reduction in detecting process.

To overcome these problems, we propose a highly efficient and effective method EK-Net which has three-stages. Firstly, images are fed to the convolutional neural network to predict the kernel map, the threshold map and the expand distance map, their schematic diagram are demonstrated in Fig.2. Secondly, the kernel map and the threshold map are combined to form the text instance roughly. Lastly, the text instance is expanded completely outwards using the expand distance map. 

The main contributions of this paper are as follows:
\begin{itemize}
\item{We suggest a new perspective EK-Net to efficiently complete the text detection of arbitrary shapes, avoiding \textquotedblleft shrinked kernel\textquotedblright 
issue generated in detection progress.}
\item{We propose the expand distance map to compensate for the previous deficiency which is more robust, making the generation of text contours more accurate.}
\item{EK-Net achieves state-of-the-art or competitive results on challenging benchmarks, exhibiting a trade-off between accuracy and real-time. Evaluation results and studies provide valuable insights into its properties.}
\end{itemize}

\begin{figure*}[t]
    \centering
        \includegraphics[width=2\columnwidth]{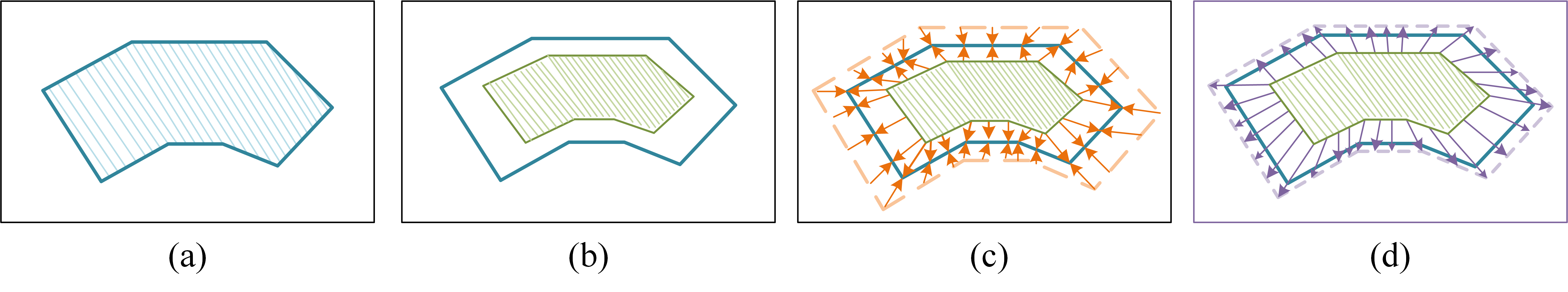}
      \caption{Label generation. (a) Text instance; (b) Text kernel; (c) Text kernel threshold; (d) Text expand distance.}\label{fig:F2}
\end{figure*}

\section{Method}
\label{sec:method}
In this section, we first introduce the overall architecture of the network in Section 2.1. Section 2.2 shows the generation process of the ground truth at three-stages. Lastly, we elaborate on training details and give the loss function.
\subsection{Overview}
\label{ssec:overview}
As discussed before, existing methods either can not effectively capture the contours of irregularly shaped text or introduce the \textquotedblleft shrinked kernel\textquotedblright issue which leads to a decrease in accuracy. To achieve simultaneous high detection accuracy and speed for arbitrary-shaped text instances, we construct EK-Net which using three-stages combined strategy.

EK-Net is a novel anchor-free text detector and the procedure of text contour reconstruction is shown in Fig.3. It is consisted of backbone, feature fusion, and post- processing. To ensure the trade-off of accuracy and real-time performance, we choose Res-Net18 as the backbone instead of Res-Net50. Feature fusion adopt FPN structure to integrate global and local information from backbone. The feature extraction network finally generates five feature maps for subsequent three-stages post-processing. The mathematical explanation is as follows.

Given an image \textit{I}, its text instances ground-truth are denoted as $ \left\{\ T_1, T_2, ..., T_i, ...\right\} $, where $T_i$ represents the \textit{i}-th text instance (blue shaded areas in Fig. 2(a)). Firstly, each text instance $T_i$ corresponds to a text kernel $K_i$, which is a shrinked version of the original text region. And $K_i$ is a subset of its corresponding text instance $K_i \subseteq T_i$, setting a scaling factor $R \in (0,1)$, $K_i = T_i \times R$. All components combined methods treat $K_i$ as the primary basic of the pixel aggregation. Secondly, the corresponding threshold $T\_S_i$ (as shown in Fig. 2(c)) of each text instance $T_i$ includes the inner and the outer parts of text contour, $T\_S_i = T_i \times R'-K_i, R' \in (1,2)$. Specially, the kernel map and the threshold map are combined to form the text instance. Lastly, the text instance kernel is expanded outwards using the expand distance map defined $E_i$. By performing two steps dilation on text instance $T_i$, we obtain $expand1\_T_i$ and $expand2\_T_i$. Subtracting $expand1\_T_i$ from $expand2\_T_i$, we get all surrounding pixels of $T_i$. $E_i$ stores all these pixels' distance to their nearest $K_i$. 

\begin{figure}[t]
    \centering
        \includegraphics[width=1.0\columnwidth]{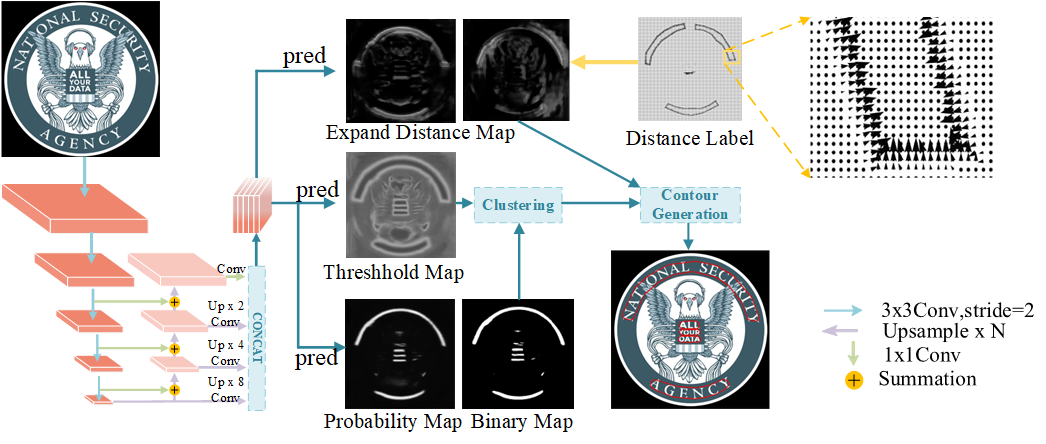}
      \caption{Overview of the proposed EK-Net. Given an input image to feature extraction net, red rectangles represent the backbone and pink rectangles are feature fusion components of FPN. Three prediction arrows followed by three-stages regression maps. }\label{fig:F3}
\end{figure}

\subsection{Label Generation}
\label{ssec:label}
The label generation process corresponds to the three-stages combined strategy as well. Firstly, the kernel map $K_i$ (green shaded areas in Fig. 2(b)) is generated by PSENet \cite{Li2018ShapeRT}, which shrinking the polygon by using the Vatti clipping algorithm \cite{Vatti1992AGS}. The scaling factor $R$ of shrinking is computed as:
\begin{equation}\label{eq:R}
  R= \frac{A\times(1-r^2)}{L},
\end{equation}
where $r$ is the shrink ratio, set to 0.4 empirically, $L$ is the perimeter and $A$ is the area of the original polygon. 

The threshold map is generated according to DB-Net \cite{liao2020real}, the text instance polygon is defined as $G$, and $G$ is dilated with the same offset $R$ to $G_d$. We consider the gap between $K_i$ and $G_d$ as the border of the text regions, where the label of the threshold map $T\_S_i$ can be generated by computing the distance to the closest segment in $G$.  

As for the expand distance $E_i$ shown in Fig. 2(d), we get all surrounding pixels through $expand1\_T_i$ and $expand2\_T_i$, then performing their polygon expansion to obtain $P$. The expand distance map label is conveyed by calculating the vector from pixel in $P$ to the nearest $K_i$.

\subsection{Optimization}
\label{ssec:opti}
The loss function $L$ is expressed as a weighted sum of the loss for the probability map of kernel $L_k$, the loss for the threshold map $L_t$, and the loss for the expand distance map $L_e$:
\begin{equation}\label{eq:L}
  L= \alpha L_k + \beta L_t + \gamma L_e,
\end{equation}
where $\alpha, \beta, \gamma$ are hyperparameters to balance the weights of the three losses. According to the numeric values of the losses, $\alpha, \beta$, and $\gamma$ are set to 1.0, 10.0, 4.0 respectively.

The probability map of kernel $L_k$ is defined as $L_{kp}$, and we apply a binary cross-entropy (BCE) loss for it which is:

\begin{equation}\label{eq:bce}
  L_{kp} = \sum_{i\in S_l}\hat y_i \log{y_i} + (1-\hat y_i)\log(1-y_i),
\end{equation}
where  $y_i$ and $\hat y_i$ respectively represent the label value and predicted value of pixel $i$, $S_l$ is the sampled set where the ratio of positives and negatives is $1 : 3$. Then we find that training with the binary of $L_k$ can eliminate noise interference and balance the Precision and Recall of output to improve the accuracy which is shown in EXPERIMENTS. We define binary map as $\hat B$:
\begin{equation}\label{eq:B}
  \hat B_{i,j} = \frac 1 {1-e^{-k(P_{i,j}-T \_ S_{i,j})}},
\end{equation}
where $T \_ S_{i,j}$ is the threshold map learned from network and  $k$ indicates the amplifying factor which set 50. The binary loss of $L_k$ is defined as $L_{kb}$, dice\_loss is employed for it: 

\begin{equation}\label{eq:Dice}
  L_{kb} =1- \frac {2\sum_{i=1}^{N}y_i \hat y_i} {{\sum_{i=1}^{N}y_i}+{\sum_{i=1}^{N}\hat y_i}},
\end{equation}
where $N$ represents the total number of pixels, which is equal to the number of pixels in a single image multiplied by the batch size.

$L_t$ is calculated as the sum of $L_1$ distance between the predicted and labeled values within the dilated text polygon $G_d$. $L_e$ is computed as the sum of smooth $L_1$ distance and $x_i = \hat y_i -y_i$.
\begin{equation}\label{eq:L1}
  L_t =\sum_{i\in R_d}\vert x_i \vert
\end{equation}

\begin{equation}
L_e =\left\{\begin{array}{l}
\begin{aligned}
0.5 \mathrm{x_i}^{2} \quad |\mathrm{x_i}|<1 \\
|\mathrm{x_i}|-0.5 \quad |\mathrm{x_i}| \geq 1
\end{aligned}
\end{array}\right.
\end{equation}

\section{Experiments}
\label{sec:exp}
To verify the efficiency and effectiveness of the proposed method, we conduct extensive experiments compared with other state-of-the-art approaches on two challenging public benchmarks. ICDAR 2015 \cite{karatzas2015icdar} represents street-scene datasets focusing on oriented text, which has considerable aspect ratios. It consists of 1000 training images and 500 testing images, text instances are labeled at the word level. CTW1500 \cite{liu2019curved} represents curved text datasets and consists of 1000 training images and 500 testing images. The text instances are annotated at the text-line level.

\subsection{Implementation Details and Metrics}
To show the advantages of the model proposed, we present several commonly used metrics which are divided into three categories. \emph{1)} Precision, Recall, and F-measure(“P”, “R”, and “F”) exhibit the model performance indicator. F-measure is a comprehensive evaluation of Precision and Recall, so we consider it as the most important metric for accuracy. \emph{2)} Param stands for the amount of parameters computation. \emph{3)} FPS is the real-time performance indicator. FPS represents the number of pictures model can process in one second. Our research is conducted on a single Nvidia GeForce RTX 2080Ti. The Adam optimizer is employed with a learning rate of 0.001.
\begin{table}[t]
    \centering
     \caption{Ablation study about three-stages regression on ICDAR 2015.}\label{tab:data1}
    \scalebox{1}{
    \begin{tabular}{cccc}
      \toprule 
        \bfseries Loss & \bfseries F  & \bfseries P & \bfseries R  \\
      \midrule 
     $L_k(L_{kp}+L_{kb})$ + $L_t$ + $L_e$   & \bfseries 85.72 & 92.00  & \bfseries 80.24  \\
     $L_k(L_{kp})$ + $L_t$ + $L_e$    & 85.10 & \bfseries93.05 & 78.40  \\ 
     $L_k(L_{kp}+L_{kb})$ + $L_t$     & 82.25 & 86.75 & 78.42  \\
      \bottomrule 
    \end{tabular}}
\end{table}

\subsection{Ablation Study about Three-Stages Regression. }
We conduct an ablation study on the  ICDAR 2015 \cite{karatzas2015icdar} dataset to show the significance of our proposed three-stages regression. The detailed experimental results are shown in Table \ref{tab:data1}. Based on the results presented in the first and second rows, we can draw the following conclusions. While the influence of $L_{kb}$ on the final results is relatively minimal in terms of F-measure, the absence of $L_{kb}$ significantly disrupts the balance between Precision and Recall. Consequently, we introduce dice\_loss which calculates the overlap between predicted and true results, and optimizes the model by minimizing the disparity between them. The incorporation of dice\_loss not only establishes equilibrium between Precision and Recall but also attains the optimal F-measure. The comparison between the first and third rows demonstrates the effectiveness of our proposed text expand distance. By incorporating the distance between the external pixels of the text and the kernel during training, we achieve a forced expansion of the text instance.

\subsection{Comparison with State-of-the-Art Methods}
\begin{figure}[htb]
    \centering
        \includegraphics[width=1.0\columnwidth]{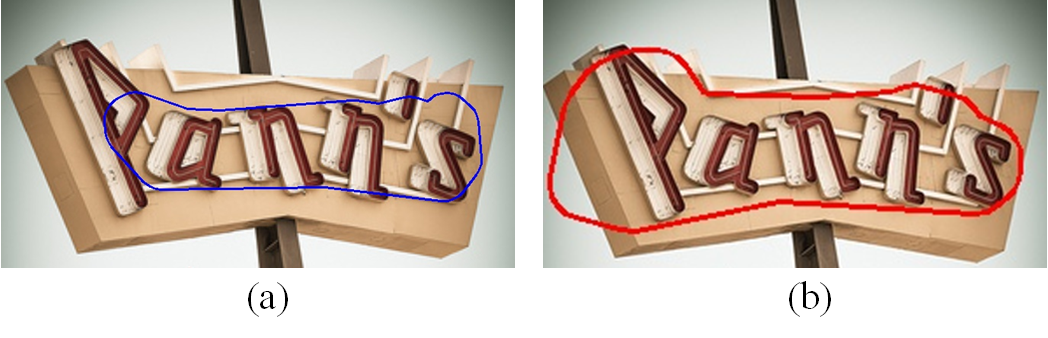}
      \caption{Visual results of previous methods (a) and EK-Net (b).}\label{fig:F4}
\end{figure}

 We evaluate our EK-Net on ICDAR 2015 \cite{karatzas2015icdar} and CTW1500 \cite{liu2019curved} to verify the superiority of our method. The results of the comparison between State-of-the-Art methods are presented in Table \ref{tab:data2} and Table \ref{tab:data3}. “*” indicates reproducing results in same environment. In terms of FPS, we achieve the fastest inference speed on both datasets. From the perspective of feature extraction methods and model size, most models adopt Res50 which size is nearly 26M. Due to the fusion of multi-channel feature maps, these methods' overall sizes surpass 26M and are close to 30M. The size of the model using Res-Net18 as the backbone(PAN, DB-Net, DB-Net++) is basically the same. Obviously, the response speed of these models is faster than that of Res-Net50. In particular, R-Net reaches a maximum size of 50.09M, making it 3.97 times larger than ours. From the results of  Table \ref{tab:data2}, DB-Net++(Res50) achieves the highest F-measure,  but DB-Net++(Res18) which is at the same level as our model, achieves a lower performance of 83.1 compared to our 85.72. In addition, EK-Net surpasses the performance of all same level models, even including those of larger sizes such as DB-Net(Res50), R-Net and Keserwani et al.. Besides, we achieve the highest Precision and Recall is also competitive among all models. The results on CTW1500 \cite{liu2019curved} are better than those on ICDAR 2015 \cite{karatzas2015icdar}. We achieve the best F-measure and Recall which surpass those of each method including DB-Net++(Res50). All these indicate that EK-Net obtains state-of-the-art or competitive results and achieves a trade-off between accuracy and real-time. The visualizations results between previous limitation methods and EK-Net are shown in Fig\ref{fig:F4}, which indicating EK-Net solve the \textquotedblleft shrinked kernel\textquotedblright issue.
 
\begin{table}[t]
    \centering
     \caption{Results of text detection on ICDAR 2015.}\label{tab:data2}
    \scalebox{0.7}{
    \begin{tabular}{ccccccc}
      \toprule 
        \bfseries Method(Backbone) & \bfseries Paper  & \bfseries F & \bfseries P &\bfseries R&\bfseries FPS& \bfseries Param  \\
      \midrule 
      PAN(Res18)  & ICCV'19 & 82.9 & 84.0 & 81.9 & 26.1 &- \\
      DB-Net(Res18)  & AAAI'20 & 82.3 & 86.8 & 78.4 &  33.51* &12.94M \\
      DB-Net(Res50)  &  AAAI'20 & 85.4 & 88.2 & 82.7 & 26& 26M$\uparrow$  \\
       PAN++(Res18)  &  TPAMI'21 & 83.1 & 85.9 & 80.4 & 28.20 & -  \\
      R-Net(VGG16)  & TMM'21 & 85.6 & 88.7 &  82.8 & 21.4 & 50.09M \\
      Keserwani et al.(Res50)  &  TCSVT'22 & 84.3 & 85.9 & 82.7 & - & 26M$\uparrow$  \\
      DB-Net++(Res18)  &  TPAMI’22 & 83.1 & 90.1 & 77.2 & 25.38* & 13M$\uparrow$  \\
      DB-Net++(Res50)  &  TPAMI’22 &  \bfseries 87.3 &90.9 &  \bfseries 83.9 & 10& 26M$\uparrow$  \\
      BIP-Net(-)  &  ICASSP'22 & 83.9 & 86.9 & 82.1 & 24.8 & - \\
      \midrule 
      Ours(Res18)    &-    & 85.72 & \bfseries 92.00  &  80.24 & \bfseries35.42 &\bfseries 12.81M  \\
      \bottomrule 
    \end{tabular}}
\end{table}

\begin{table}[t]
    \centering
     \caption{Results of text detection on CTW1500.}\label{tab:data3}
    \scalebox{0.7}{
    \begin{tabular}{ccccccc}
      \toprule 
        \bfseries Method(Backbone) & \bfseries Paper  & \bfseries F & \bfseries P &\bfseries R&\bfseries FPS& \bfseries Param  \\
      \midrule 
      PAN(Res18)  & ICCV'19 & 83.7 & 86.4 & 81.2 &  39.8 &- \\
      DB-Net(Res18)  & AAAI'20 & 81.0 & 84.8 & 77.5 & 37.41* & 12.94M \\
      DB-Net(Res50)  &  AAAI'20 & 83.4 & 86.9 & 80.2 & 22& 26M$\uparrow$  \\
       PAN++(Res18)  &  TPAMI'21 & 84.0 & 87.1 & 81.1 & 36.0 & -  \\
      OPMP(Res50)  & TMM'21 & 82.9 & 85.1 & 80.8 & 1.4 &26M$\uparrow$  \\
      TextBPN(Res50)  & ICCV'21 & 84.5 &87.8 & 81.4 & 12.1 &26M$\uparrow$  \\
      zhao et al.(Res50)  &  TIP'22 & 84.1 & 86.1 & 82.1 & - & 26M$\uparrow$  \\
      DB-Net++(Res18)  &  TPAMI’22 & 83.9 & 86.7 & 81.3 & 27.73* &13M$\uparrow$  \\
      DB-Net++(Res50)  &  TPAMI’22 & 85.1 & \bfseries88.5 & 82.0 & 21 &26M$\uparrow$  \\
      BIP-Net(-)  &  ICASSP'22 & 85.1 & 87.7 &  82.6 & 35.6 & - \\
      \midrule 
      Ours(Res18)    &-    & \bfseries 85.75 &  87.85  & \bfseries 83.74 &\bfseries40.13  & \bfseries 12.81M \\
      \bottomrule 
    \end{tabular}}
\end{table}

\section{Conclusions}
\label{sec:con}
In this paper,  we find a special issue that has never been noticed, which leads to performance degradation. Then we propose a novel framework for detecting arbitrary-shaped scene text, which includes three-stages regression to complete text instance detection. The experiments verify that our method outperforms the state-of-the-art methods on standard scene text benchmarks, in terms of speed and accuracy. In the future, we are keen on expanding our method to encompass end-to-end text spotting for efficient text-related applications.\\

\bibliographystyle{IEEEbib}
\bibliography{strings,small_refs}

\begin{thebibliography}{10}

\bibitem{Zhang2021PointerNF}
Yi~Zhang, Wei Yang, et~al.,
\newblock ``Pointer networks for arbitrary-shaped text spotting,''
\newblock {\em ICASSP 2021 - 2021 IEEE International Conference on Acoustics,
  Speech and Signal Processing (ICASSP)}, pp. 2375--2379, 2021.

\bibitem{Zhou2016SemanticUO}
Bolei Zhou et~al.,
\newblock ``Semantic understanding of scenes through the ade20k dataset,''
\newblock {\em International Journal of Computer Vision}, vol. 127, pp. 302 --
  321, 2016.

\bibitem{Long2018SceneTD}
Shangbang~Long He et~al.,
\newblock ``Scene text detection and recognition: The deep learning era,''
\newblock {\em International Journal of Computer Vision}, vol. 129, pp. 161 --
  184, 2018.

\bibitem{Wang2021TowardsRV}
Jiapeng Wang, Chongyu Liu, et~al.,
\newblock ``Towards robust visual information extraction in real world: New
  dataset and novel solution,''
\newblock {\em ArXiv}, vol. abs/2102.06732, 2021.

\bibitem{Jaderberg2014ReadingTI}
Max Jaderberg, Karen Simonyan, Andrea Vedaldi, and Andrew Zisserman,
\newblock ``Reading text in the wild with convolutional neural networks,''
\newblock {\em International Journal of Computer Vision}, vol. 116, pp. 1--20,
  2014.

\bibitem{He2016IdentityMI}
Kaiming He, X.~Zhang, Shaoqing Ren, and Jian Sun,
\newblock ``Identity mappings in deep residual networks,''
\newblock in {\em European Conference on Computer Vision}, 2016.

\bibitem{Li2018ShapeRT}
Xiang Li, Wenhai Wang, et~al.,
\newblock ``Shape robust text detection with progressive scale expansion
  network,''
\newblock {\em 2019 IEEE/CVF Conference on Computer Vision and Pattern
  Recognition (CVPR)}, pp. 9328--9337, 2018.

\bibitem{zhu2021fourier}
Yiqin Zhu, Jianyong Chen, et~al.,
\newblock ``Fourier contour embedding for arbitrary-shaped text detection,''
\newblock in {\em Proceedings of the IEEE/CVF conference on computer vision and
  pattern recognition}, 2021, pp. 3123--3131.

\bibitem{dai2021progressive}
Pengwen Dai, Sanyi Zhang, Hua Zhang, and Xiaochun Cao,
\newblock ``Progressive contour regression for arbitrary-shape scene text
  detection,''
\newblock in {\em Proceedings of the IEEE/CVF conference on computer vision and
  pattern recognition}, 2021, pp. 7393--7402.

\bibitem{Liao2016TextBoxesAF}
Minghui Liao, Baoguang Shi, Xiang Bai, Xinggang Wang, and Wenyu Liu,
\newblock ``Textboxes: A fast text detector with a single deep neural
  network,''
\newblock {\em ArXiv}, vol. abs/1611.06779, 2016.

\bibitem{Liu2015SSDSS}
W.~Liu, Dragomir Anguelov, D.~Erhan, Christian Szegedy, Scott~E. Reed,
  Cheng-Yang Fu, and Alexander~C. Berg,
\newblock ``Ssd: Single shot multibox detector,''
\newblock in {\em European Conference on Computer Vision}, 2015.

\bibitem{liao2018textboxes++}
Minghui Liao, Baoguang Shi, and Xiang Bai,
\newblock ``Textboxes++: A single-shot oriented scene text detector,''
\newblock {\em IEEE transactions on image processing}, vol. 27, no. 8, pp.
  3676--3690, 2018.

\bibitem{liu2017deep}
Yuliang Liu and Lianwen Jin,
\newblock ``Deep matching prior network: Toward tighter multi-oriented text
  detection,''
\newblock in {\em Proceedings of the IEEE conference on computer vision and
  pattern recognition}, 2017, pp. 1962--1969.

\bibitem{xie2019scene}
Enze Xie, Yuhang Zang, et~al.,
\newblock ``Scene text detection with supervised pyramid context network,''
\newblock in {\em Proceedings of the AAAI conference on artificial
  intelligence}, 2019, vol.~33, pp. 9038--9045.

\bibitem{he2017mask}
Kaiming He, Gkioxari, et~al.,
\newblock ``Mask r-cnn,''
\newblock in {\em Proceedings of the IEEE international conference on computer
  vision}, 2017, pp. 2961--2969.

\bibitem{Vatti1992AGS}
Bala~R. Vatti,
\newblock ``A generic solution to polygon clipping,''
\newblock {\em Commun. ACM}, vol. 35, pp. 56--63, 1992.

\bibitem{liao2020real}
Minghui Liao, Zhaoyi Wan, Cong Yao, Kai Chen, and Xiang Bai,
\newblock ``Real-time scene text detection with differentiable binarization,''
\newblock in {\em Proceedings of the AAAI conference on artificial
  intelligence}, 2020, vol.~34, pp. 11474--11481.

\bibitem{Liao2019RealTimeST}
Minghui Liao, Zhaoyi Wan, et~al.,
\newblock ``Real-time scene text detection with differentiable binarization and
  adaptive scale fusion,''
\newblock {\em IEEE Transactions on Pattern Analysis and Machine Intelligence},
  vol. 45, pp. 919--931, 2019.

\bibitem{Wang2019EfficientAA}
Wenhai Wang, Enze Xie, et~al.,
\newblock ``Efficient and accurate arbitrary-shaped text detection with pixel
  aggregation network,''
\newblock {\em 2019 IEEE/CVF International Conference on Computer Vision
  (ICCV)}, pp. 8439--8448, 2019.

\bibitem{wang2021pan++}
Wenhai Wang, Enze Xie, Xiang Li, et~al.,
\newblock ``Pan++: Towards efficient and accurate end-to-end spotting of
  arbitrarily-shaped text,''
\newblock {\em IEEE Transactions on Pattern Analysis and Machine Intelligence},
  vol. 44, no. 9, pp. 5349--5367, 2021.

\bibitem{sheng2021centripetaltext}
Tao Sheng, Jie Chen, and Zhouhui Lian,
\newblock ``Centripetaltext: An efficient text instance representation for
  scene text detection,''
\newblock {\em Advances in Neural Information Processing Systems}, vol. 34, pp.
  335--346, 2021.

\bibitem{Yang2021BiPNetBP}
Chuan Yang, Mulin Chen, et~al.,
\newblock ``Bip-net: Bidirectional perspective strategy based arbitrary-shaped
  text detection network,''
\newblock {\em ICASSP 2022 - 2022 IEEE International Conference on Acoustics,
  Speech and Signal Processing (ICASSP)}, pp. 2255--2259, 2021.

\bibitem{karatzas2015icdar}
Dimosthenis Karatzas, Lluis Gomez-Bigorda, et~al.,
\newblock ``Icdar 2015 competition on robust reading,''
\newblock in {\em 2015 13th international conference on document analysis and
  recognition (ICDAR)}. IEEE, 2015, pp. 1156--1160.

\bibitem{liu2019curved}
Yuliang Liu, Lianwen Jin, Shuaitao Zhang, Canjie Luo, and Sheng Zhang,
\newblock ``Curved scene text detection via transverse and longitudinal
  sequence connection,''
\newblock {\em Pattern Recognition}, vol. 90, pp. 337--345, 2019.

\end{thebibliography}

\end{document}